\documentclass[5p]{elsarticle}

\usepackage{amsmath,amsfonts}
\DeclareMathOperator*{\argmin}{arg\,min}
\usepackage{url}
\usepackage{subcaption}
\usepackage[hidelinks]{hyperref}

\journal{Neural Networks}

\begin{document}

\begin{frontmatter}

\title{Single-Layer Vision Transformers for More Accurate Early Exits with Less Overhead}

\author[inst1]{Arian~Bakhtiarnia\corref{cor1}}
\ead{arianbakh@ece.au.dk}
\author[inst1]{Qi~Zhang}
\ead{qz@ece.au.dk}
\author[inst1]{Alexandros~Iosifidis}
\ead{ai@ece.au.dk}

\affiliation[inst1]{organization={DIGIT, Department of Electrical and Computer Engineering, Aarhus University},
            addressline={Finlandsgade 22}, 
            city={Aarhus},
            postcode={8200}, 
            state={Midtjylland},
            country={Denmark}}

\begin{abstract}
Deploying deep learning models in time-critical applications with limited computational resources, for instance in edge computing systems and IoT networks, is a challenging task that often relies on dynamic inference methods such as early exiting. In this paper, we introduce a novel architecture for early exiting based on the vision transformer architecture, as well as a fine-tuning strategy that significantly increase the accuracy of early exit branches compared to conventional approaches while introducing less overhead. Through extensive experiments on image and audio classification as well as audiovisual crowd counting, we show that our method works for both classification and regression problems, and in both single- and multi-modal settings. Additionally, we introduce a novel method for integrating audio and visual modalities within early exits in audiovisual data analysis, that can lead to a more fine-grained dynamic inference.
\end{abstract}


\cortext[cor1]{Corresponding author}

\end{frontmatter}


\section{Introduction}

Over the past decade, deep learning has shown tremendous success across various fields, such as computer vision and natural language processing \cite{LeCun2015}. However, deep learning models are by definition composed of many layers of interconnected neurons, even reaching billions of parameters, which makes them computationally expensive. This has sparked a great deal of research in order to make deep learning models more lightweight, for which many approaches have been proposed, for instance, \textit{model compression} methods \cite{Cheng2018} such as \textit{quantization} \cite{Rastegari2016}, \textit{pruning} \cite{1608.08710}, \textit{low-rank approximation} \cite{tran2018improving} and \textit{knowledge distillation} \cite{1503.02531}.

More and more emerging internet of things (IoT) applications are integrating deep learning models, such as video surveillance, voice assistants, augmented reality and cooperative autonomous driving, which are often time-sensitive and require inputs to be processed within specific deadlines \cite{Chen2019, Wang2020}. The heavy computational burden of deep learning becomes problematic for these time-critical IoT applications, due to resource-constrained IoT devices. \textit{Edge computing} is a promising computing paradigm for addressing this issue, in which the deep learning task is offloaded to edge servers in the proximity of IoT devices.

Since edge computing systems introduce computation offloading over a communication network and involve multiple nodes working collaboratively in order to complete the task in a timely manner, transmission time has to be taken into account in addition to the deep learning computation time. However, transmission time may vary greatly over time and across different channels. Consequently, deep learning models running on edge computing systems and IoT networks should be capable of \textit{anytime prediction}, meaning they should be able to provide a valid response even if they are interrupted before traversing the entire neural network, although the model is expected to provide a better answer if it is allowed to run for longer time.

\textit{Dynamic inference} approaches \cite{2102.04906} modify the computation graph based on each input during the inference phase in order to fit the time constraints. A dynamic inference approach that particularly suits anytime prediction is \textit{early exiting} \cite{Scardapane2020}, also referred to as \textit{multi-exit architectures} or \textit{auxiliary classifiers} in the literature. In multi-exit architectures, one or more early exit \textit{branches} are placed after some of the intermediate hidden layers of the \textit{backbone} network. The goal of each of these branches is to provide an early result similar to the final result of the neural network using only the features extracted up to that particular branch location. These early results are inevitably less accurate than the final result of the network. In order to achieve anytime prediction using early exiting, the latest early result can be used whenever the execution is interrupted, for instance, whenever a hard deadline is reached. Computation time can be further decreased by applying model compression techniques on the backbone of multi-exit architectures. Besides anytime prediction, early exiting can also be used in \textit{budgeted batch classification} where a fixed amount of time is available in order to classify a set of input samples. In such a setting, the result of earlier branches can be used for ``easier'' samples whereas the result of later branches or the final result can be used for ``harder'' ones. The difficulty of each sample can be determined based on the confidence of the network about its output \cite{Teerapittayanon2016}, although other approaches exist in the literature \cite{Scardapane2020}.

Early exit branches are expected to have a low overhead in terms of the extra computation they introduce, since a high overhead would defeat the purpose. Therefore, they often contain only a handful of layers. Ideally, we want the accuracy of the early results to be close to that of the final result, since a higher accuracy for early exit branches means that the overall reliability of the system increases. However, the low-overhead constraint makes it quite challenging to achieve a high accuracy since the early exit branches have significantly less trainable parameters compared to the rest of the network. Several approaches for increasing the accuracy of early exits such as knowledge distillation \cite{Phuong2019}, curriculum learning \cite{2104.10461} and architectures designed specifically for early exit branches \cite{2106.15183} have been suggested. In this paper, we propose a novel architecture in order to obtain more accurate early exits for convolutional neural network (CNN) backbones.

A neural architecture called \textit{vision transformer} (\textit{ViT}) \cite{dosovitskiy2021an} has been recently introduced for image classification which is radically different from convolutional neural networks. The building blocks of Vision Transformer have been used for early exits placed on Vision Transformer backbones \cite{2106.15183}, however, using Transformer-based early exit branches on CNN backbones is not intuitive and requires additional steps and architectural modifications. We use a modified version of this architecture instead of the usual convolution and pooling layers in early exit branches and show that our method can significantly increase the accuracy of early exits compared to conventional architectures by fusing local and global receptive fields\footnote{Our code will be available at \url{https://gitlab.au.dk/maleci/sl_vit}.}. The contributions of this paper can be summarized as follows:
\begin{itemize}
  \item We propose a novel architecture for early exit branches in multi-exit architectures based on vision transformers, called \textit{single-layer vision transformer} (\textit{SL-ViT}). We compare our method with conventional CNN-based early exit architectures across 27 scenarios involving different datasets, branch locations and backbone networks and show that our method is significantly more accurate in 26 of these scenarios, while having less overhead in terms of number of parameters and floating point operators (FLOPS). To the best of our knowledge the fusion of global and local scope in early exits has never been used in multi-exit architectures before.
  \item We show that our method is a general purpose approach that works across different modalities as well as multi-modal settings by investigating image classification, audio classification and audiovisual crowd counting scenarios. We also show that our method works for both classification and regression problems.
  \item We introduce a novel way of integrating audio and visual features in early exits using vision transformers. To the best of our knowledge, this is the first time early exits have been studied in multi-modal settings.
  \item We provide insight into why our method achieves better results compared to conventional CNN-based architectures by investigating the role of attention and receptive field.
  \item We introduce a fine-tuning strategy for SL-ViT called \textit{copycat single-layer vision transformer} (\textit{CC-SL-ViT}) which is based on the copycat strategy developed for CNNs \cite{CorreiaSilva2018} and show that this method can further increase the accuracy of SL-ViT early exits. To the best of our knowledge this is the first time the copycat strategy is used for vision transformers or early exits.
\end{itemize}

The rest of this paper is organized as follows: Section \ref{S:RelatedWork} provides an overview of the relevant literature; Section \ref{S:Method} describes our proposed method in detail; Section \ref{S:ExperimentalSetup} explains the details of our experiments; Section \ref{S:Results} showcases the experiment results; and, finally, Section \ref{S:Conclusions} briefly discusses the results and concludes the paper.

\section{Related Work}\label{S:RelatedWork}

This section provides the necessary prerequisites for understanding our method and experiments. We start by describing the particulars of multi-exit architectures. Subsequently, we provide the details of the vision transformer architecture, which is the foundation of the proposed method. Then, we briefly touch on how audio classification is normally carried out, which is included in several scenarios in our experiments. Finally, we explain another scenario investigated in our experiments, i.e. crowd counting, and how it can be approached in a multi-modal manner.

\subsection{Multi-Exit Architectures}
In order to describe multi-exit architectures, we use the same notation as Scardapane et al. \cite{Scardapane2020} where a neural network is formulated as a function $f(X) = f_L(f_{L - 1}(...f_1(X))) $. In this formulation $L$ signifies the total number of layers in the network and $f_i$ is the operator corresponding to layer $i$, which can be a convolutional layer, a fully-connected layer, a normalization layer, or any other differentiable operator. $h_i = f_i(h_{i - 1})$ denotes the output of layer $i$, where $h_0$ is the input $ X $. Finally, $\theta_i$ symbolizes the trainable parameters of layer $i$.

Equation \eqref{basic_nn_loss} formulates the training process for the neural network which is achieved by tuning its parameters using an optimization algorithm on the landscape defined by a loss function. In this equation, the parameters of the neural network are denoted by $\theta = \bigcup_{i = 1}^L \theta_i$, the training samples are signified by $\{(X_n, y_n)\}_{n = 1}^N$, and $l(\cdot, \cdot)$ is the loss function.

\begin{equation}
f^* = \argmin_{\theta} \sum_{n = 1}^N l(y_n, f(X_n))
\label{basic_nn_loss}
\end{equation}

\begin{figure}[htbp]
\centerline{\includegraphics[width=0.48\textwidth]{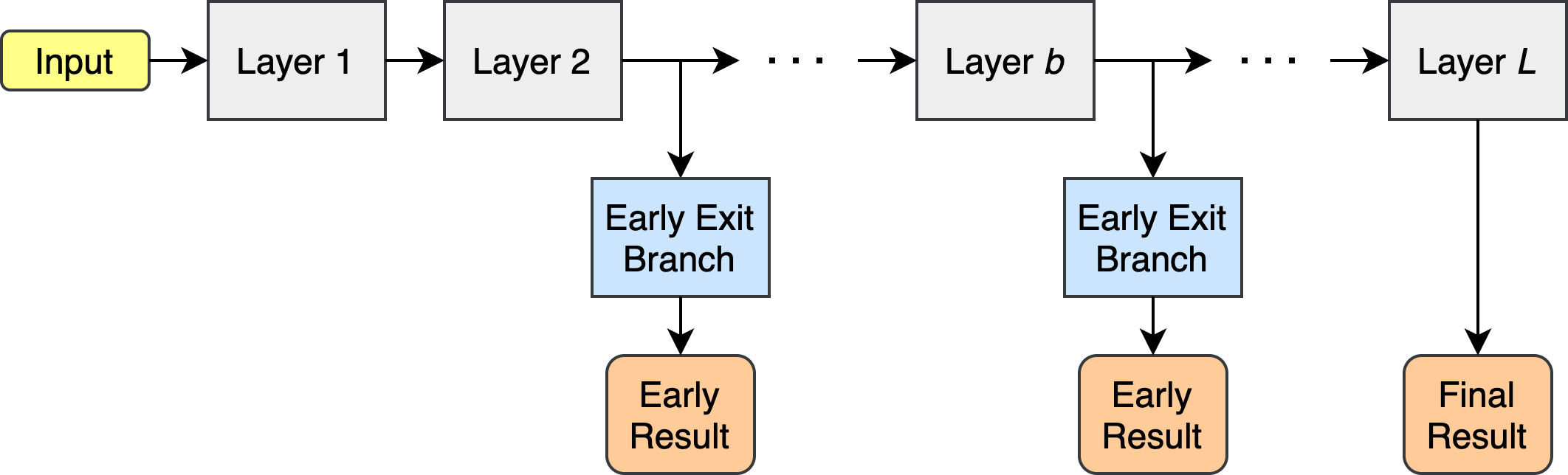}}
\caption{Schematic illustration of a multi-exit architecture with two early exits.}
\label{multi_exit}
\end{figure}

Extending this notation to multi-exit architectures, $B \subseteq \{1, .., L\}$ signifies the set of selected branch locations after which early exit branches will be placed. $c_b(h_b) = y_b $ is the classifier or regressor representing the early exit branch at each branch location $b$, where $y_b$ denotes the early result at that location. The schematic illustration of a multi-exit architecture is presented in Figure \ref{multi_exit}. However, since there are multiple outputs, and thus multiple loss signals in a multi-exit architecture, its training is not as straightforward.

Three different approaches for training multi-exit architectures exist in the literature \cite{Scardapane2020, Baccarelli2020, 2104.10461}. In the first approach, called \textit{end-to-end} training, the loss signals of all exits are combined and backpropagated through the network at the same time. With end-to-end training, the contribution of each loss signal to the total loss is expressed with weight values, which are therefore hyper-parameters of the model.
 
The second approach, called \textit{layer-wise} training, first trains the network up to and including the first exit branch. Subsequently, the part of the network that has been trained so far is frozen, meaning its parameters are not modified any further, and the remainder of the network up to and including the second exit branch is trained. This process continues until the entire network is trained. Note that with this approach, there is no guarantee that the accuracy of the final exit remains unchanged.
 
In the final approach, called \textit{classifier-wise} training, the backbone network is completely frozen and each branch is trained independent of the rest of the network and other branches, meaning the parameters $\theta$ are not modified and only the parameters of the classifers/regressors $\{c_b\}, b \in B$ are trained. With this approach, no new hyper-parameters are introduced and the backbone remains unchanged. However, the early exit branches affect a lower number of trainable parameters compared to the other approaches.

In this paper, we choose to follow the classifier-wise training approach due to its practical importance. This is because with classifier-wise training, early exit branches can be easily added on top of existing backbone networks without the need for re-training and hyper-parameter optimization, which can be computationally expensive and time consuming. Furthermore, with end-to-end and layer-wise training strategies, the number of branches and their placement can lead to further trade-offs and affect the overall performance of the model. Since branches are independently trained in the classifier-wise strategy, any number of branches can exist and a branch can be placed at any location without affecting the performance of other branches or the backbone.

It is important to mention that branches placed later in the backbone network do not necessarily result in a higher accuracy compared to branches placed earlier. The usage of such branches would therefore not be sensible since earlier branches exist that require less computation and provide more accurate results. We hereby use the term \textit{impractical} to refer to such branches.

As previously mentioned, there are several methods that try to improve the accuracy of early exits. The method in \cite{Phuong2019} uses the combination of the distillation loss from the final exit and the loss signal from ground truth labels to train more accurate early exits using in the end-to-end training setting. The method in \cite{zhang2019your} expands on this idea by adding a third loss signal based on the difference between features of the latest early exit with earlier exits. The method in \cite{li2019improved} proposes a technique called \textit{gradient equilibrium} to combat the problem of gradient imbalance that surfaces when using the end-to-end strategy, which is when the variance of the gradients becomes very large when loss signals from multiple exits are combined, leading to unstable training. Moreover, this paper introduces forward and backward knowledge transfer that aims to encourage collaboration among different exits. The method in \cite{wolczyk2021zero} improves the accuracy of later exits by reusing predictions from earlier exits. The method in \cite{jie2019anytime} circumvents the problem of impractical branches by adaptively selecting the exit location based on time budget and the specific input. The method in \cite{pomponi2021probabilistic} simplifies the design of multi-exit architectures by removing the hyper-parameters of the end-to-end training strategy that specify the contribution of each loss signal.

Besides efficient inference, early exits can prove useful in several other applications, for instance, the method in \cite{lee2021local} allows for parallel training of the segments of the DNN that exist between early exits, by training each segment based on the loss signal of the next segment obtained in the previous training stage. Moreover, early exits can be added to the network during the training in order to increase the accuracy of the backbone network and discarded after the training phase, for instance, the widely used Inception model \cite{szegedy2015going} was trained in this way.

Besides early exiting, several other approaches exist for dynamic inference, for instance, layer skipping \cite{Elbayad2020Depth-Adaptive, 1603.08983, 2107.05407, Wang_2018_ECCV} where the execution of some of the layers of the DNN are skipped, and channel skipping \cite{GaoZDMX19} where less impactful channels of convolutional neural networks are ignored and their computation is skipped during the inference phase. However, unlike early exits, these approaches cannot provide an output if the execution is interrupted due to a strict deadline, as these methods need to perform the computations until the very last layer.

\subsection{Vision Transformer}\label{S:RelatedWork_ViT}
The transformer architecture was first introduced by Vaswani et al. \cite{NIPS2017_3f5ee243} for natural language processing, and it has recently been adapted for solving computer vision problems by Dosovitskiy et al. \cite{dosovitskiy2021an}. Vision transformer was originally developed for the problem of image classification, however, variations of vision transformer have since been applied to many computer vision problems, such as object detection, depth estimation, semantic segmentation, image generation and action recognition, as well as multi-modal data analysis tasks such as text-to-image synthesis and visual question answering \cite{2012.12556, 2101.01169, 2011.14141}.

In order to describe the vision transformer architecture, we first explain the \textit{self-attention} layer. The input of this layer is in the form of a sequence $ X = (x_1, \dots, x_n) $ where $ X \in \mathbb{R}^{n \times d} $ and $ d $ is the embedding dimension to represent each entity. Its output is in the form of $ Z = (z_1, \dots, z_n) $ where $ Z \in \mathbb{R}^{n \times d_v} $. The goal of self-attention is to capture the interaction between the entities in the sequence. For this purpose, each vector $ x_i $ in the sequence is transformed into three separate vectors: the \textit{query} vector $ q_i \in \mathbb{R}^{d_q} $, the \textit{key} vector $ k_i \in \mathbb{R}^{d_k} $ and the \textit{value} vector $ v_i \in \mathbb{R}^{d_v} $, where $ d_q = d_k $. To construct the output vector $ z_i $ that corresponds to the input $ x_i $, for each vector $ x_j $ in $ X $ (including $ x_i $ itself), the scalar $ a_{ij} $ is calculated by the inner product of $ q_i $ and $ k_j $. Output vector $ z_i $ is then calculated by summing the value vectors $ v_1, \dots, v_n $ weighted by their corresponding scalars, that is, $ z_i = \sum_{j = 1}^n a_{ij} v_j $. The scalar $ a_{ij} $ basically specifies how much attention the $ i $-th entity should pay to the $ j $-th entity, since $ a_{ij} $ determines the contribution of $ v_j $ to the combined output $ z_i $. In practice, the scalars are normalized by $ \sqrt{d_k} $ and converted into probabilities using the softmax function.

If the key, query and value vectors are packed into matrices $ Q = XW^Q $, $ K = XW^K $ and $ V = XW^V $, where $ W^Q $, $ W^K $ and $ W^V $ are learnable weight matrices, the above operation can be rephrased as follows:
\begin{equation}
Z = \mathit{softmax} \left(\frac{QK^T}{\sqrt{d_k}} \right)V
\label{self_attention}
\end{equation}

In order to enable the model to capture more than one type of relationship between the entities in the sequence, self-attention is extended to \textit{multi-head attention} by concatenating the output of $ h $ different self-attention blocks $ Z_1, \dots, Z_h $ each with its own set of learnable weight matrices, into a single matrix $ Z' = [Z_0, \dots, Z_h] \in \mathbb{R}^{n \times h . d_v} $, which is then projected using a weight matrix $ W' \in \mathbb{R}^{h . d_v \times d} $.

A \textit{transformer encoder} is constructed by passing the input sequence into a normalization layer, a multi-head attention layer, a second normalization layer and a multi-layer perceptron (MLP), respectively. Two residual connections are added, one by adding the input sequence to the output of the multi-head attention, and the other by adding the output of the multi-head attention to the output of the MLP.

\begin{figure}[htbp]
\centerline{\includegraphics[width=0.48\textwidth]{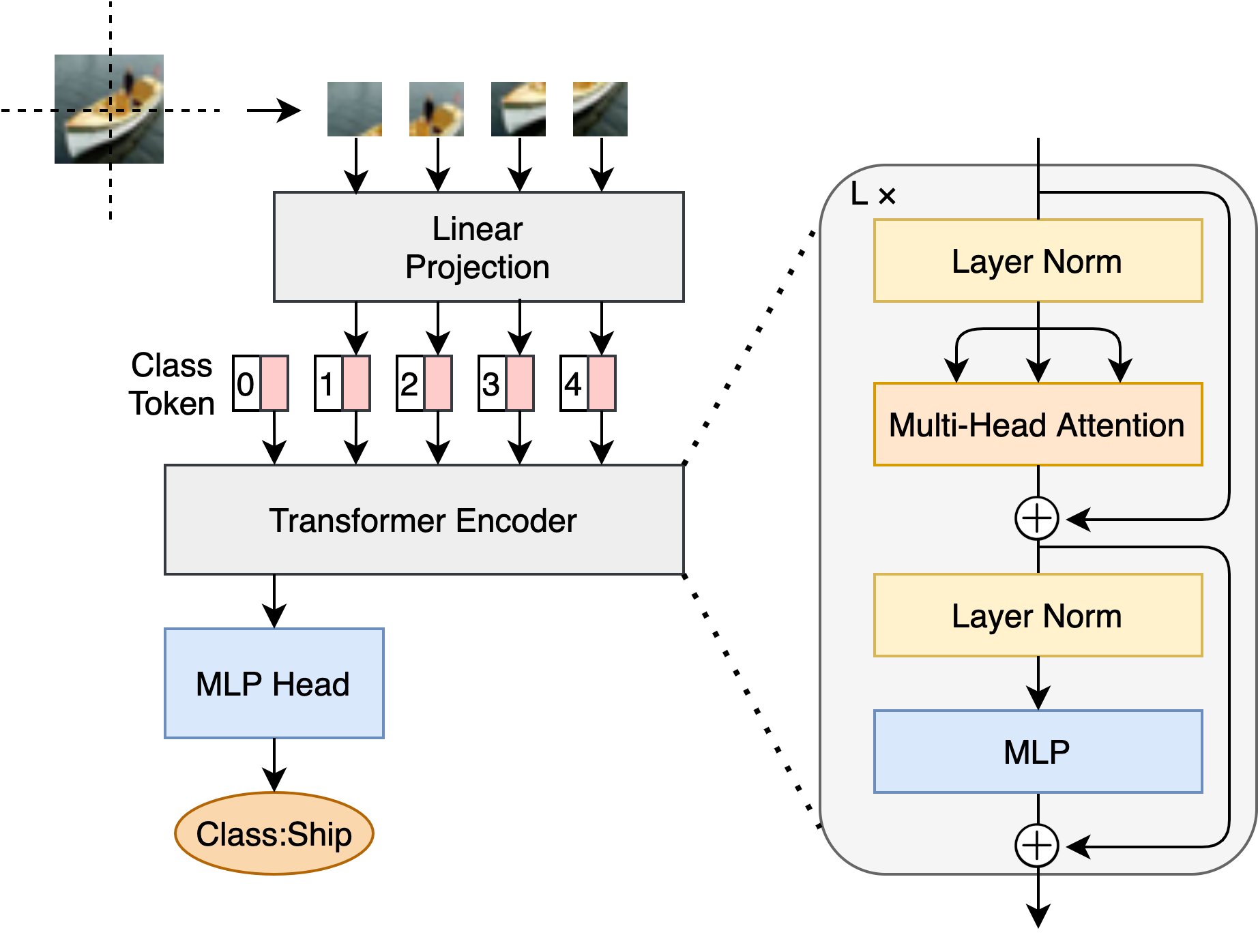}}
\caption{The vision transformer (ViT) architecture for image classification.}
\label{vit_architecture}
\end{figure}

Putting it all together, a vision transformer is created by first splitting the input image into patches. Subsequently, the sequence of patches is projected into a sequence of vectors and a positional embedding is added to the corresponding vector of each patch. An additional learnable embedding called \textit{classification token} is added to the beginning of the sequence. The sequence then passes through $ L $ transformer encoders. Finally, the first vector in the output of the last transformer encoder, which corresponds to the classification token, is passed to a MLP which outputs the final classification result. The architecture of vision transformer is depicted in Figure \ref{vit_architecture}.

ViT-EE is a method which uses transformer encoders for early exits placed on ViT backbones \cite{2106.15183}. ViT-EE uses the exact same layer as the ViT backbone. Using the building blocks of the backbone network for early exit branches is simple and intuitive, and it is the reason why so far, mostly convolutional layers have been used for early exiting CNN backbones. However, as we show in this work, carefully designing the architecture of early exit branches can lead to significant improvements. Using Transformer-based early exit branches on CNN backbones is not intuitive, and requires additional steps such as converting tensors to patches, dealing with the classification token and fine-tuning the architecture parameters including patch size, attention heads, embedding representation, the size and number of layers for MLP, and dropout. Moreover, we show that removing the last residual connection in the transformer encoder can improve the performance in some cases.

Furthermore, ViT backbones have a global receptive field in every layer, this means that ViT-EE is not necessarily ideal for early exits at all layers, as it adds too much overhead without providing improvements in terms of receptive field. On the other hand, CNN backbones have a limited receptive field particularly in earlier layers, therefore fusing this receptive field with a global one leads to improvements.

\subsection{Audio Classification}
Similar to image classification, audio classification is the problem of categorizing a given audio waveform into one of several predetermined classes. For instance, the given audio waveform could be a musical recording, and the goal could be to specify which genre of music it belongs to. 
To represent the input features, \textit{spectrograms} obtained by applying short-time Fourier transform (STFT) and \textit{Mel spectrograms} are commonly used \cite{1606.00298}, although raw audio waveforms can been used as well \cite{1712.00866}. Mel spectrograms are spectrograms that are constructed using the \textit{Mel scale} which is a nonlinear transformation of the frequency scale designed based on domain knowledge about the human auditory system.
Various deep learning models for audio classification exist in the literature, including models that are commonly used for image classification, namely ResNet \cite{He2016}, DenseNet \cite{Huang2017} and Inception \cite{Szegedy2016}, which have been shown to be quite effective for audio classification as well \cite{2007.11154}. Conveniently, the same three networks have previously been used as backbone networks when investigating early exiting for image classification \cite{2104.10461}. Therefore we use these backbone networks for both image and audio classification in our experiments.

\subsection{Audiovisual Crowd Counting}\label{S:RelatedWork_AudioCSRNet}
\textit{Crowd counting} refers to the problem of identifying the total number of people present in a given image. Crowd counting has many applications such as safety monitoring, disaster management, design of public spaces, intelligence gathering and analysis, creation of virtual environments and forensic search \cite{Sindagi2018}. With many of these applications, it is vital for the model to perform in near real-time. However, the input images in these scenarios often have high resolutions, such as HD or Full HD. Moreover, many of the available methods contain an immense number of parameters \cite{2003.12783}. This means that crowd counting models are often very computationally expensive, therefore, dynamic inference methods such as early exiting and other lightweight deep learning methods become essential in real world applications.

\begin{figure}
\centering
\begin{subfigure}{.24\textwidth}
  \centering
  \includegraphics[width=.96\linewidth]{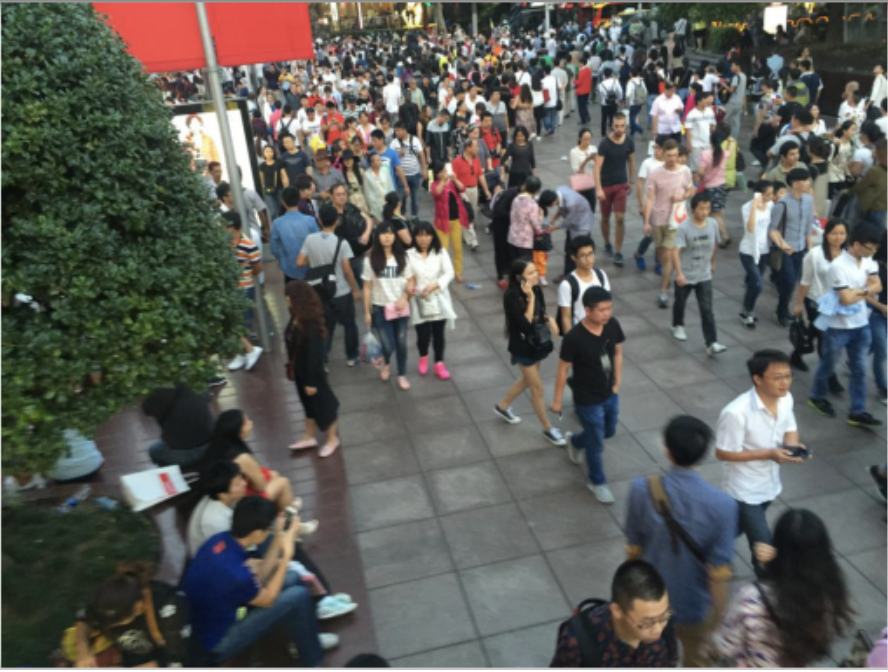}
\end{subfigure}%
\begin{subfigure}{.24\textwidth}
  \centering
  \includegraphics[width=.96\linewidth]{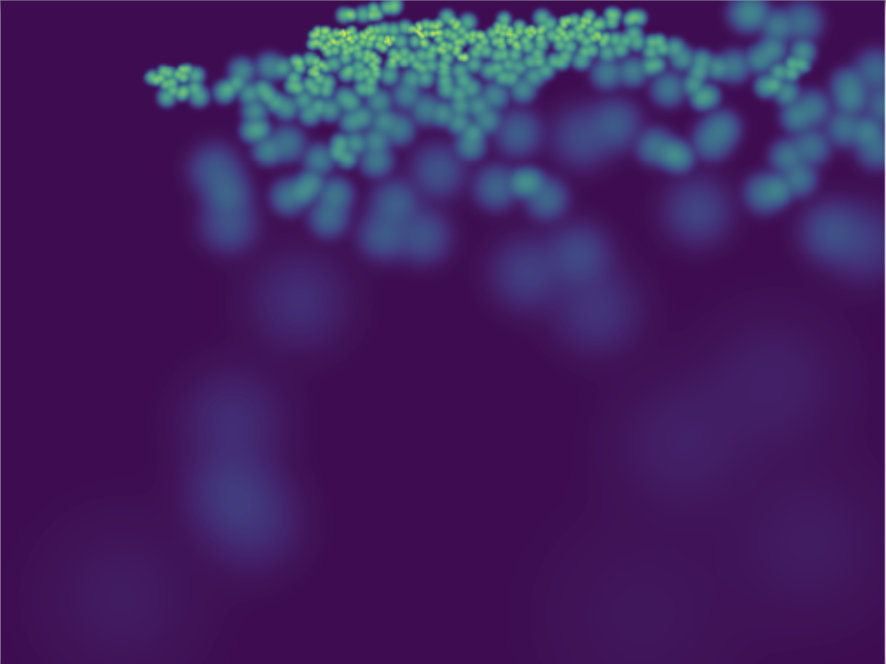}
\end{subfigure}
\caption{An example image from the Shanghai Tech dataset and its corresponding ground truth density map.}
\label{crowd_counting_label_example}
\end{figure}

Although the main objective of this task is to obtain a single count from an image, many methods treat this problem as dense prediction where the output is a \textit{density map} depicting the density of the crowd across the input image, and the total count is calculated by the sum of all values in the density map. Therefore, in most crowd counting datasets, such as Shanghai Tech \cite{Zhang2016} and World Expo '10 \cite{CongZhang2015}, the locations of the heads of individuals in the image are annotated and provided as targets. A ground truth density map can then be obtained from these \textit{head annotations} using Gaussian kernels or more complicated and specialized methods \cite{2003.12783}. Figure \ref{crowd_counting_label_example} shows an image from the Shanghai Tech dataset and the ground truth density map that was generated from the provided head annotations using the method presented in Zhang et al \cite{Zhang2016}. In crowd counting, \textit{Mean Absolute Error} (\textit{MAE}) is usually used as a measure of accuracy whereas \textit{Mean Squared Error} (\textit{MSE}) is used as a measure of robustness \cite{1906.09707}.

Many crowd counting methods exist in the literature \cite{2003.12783}, however, most of these methods are applied in a single-modal fashion where the input is an image or a video frame. In contrast, AudioCSRNet \cite{2005.07097}, a multi-modal extension of the widely-used CSRNet model for crowd counting \cite{Li2018}, takes as input the ambient audio of a scene in addition to its image. The authors show that the ambient audio improves the result in situations where the image quality is not ideal, for instance, low image resolution, presence of noise, occlusion and low illumination.

\begin{figure*}[htbp]
\centerline{\includegraphics[width=\textwidth]{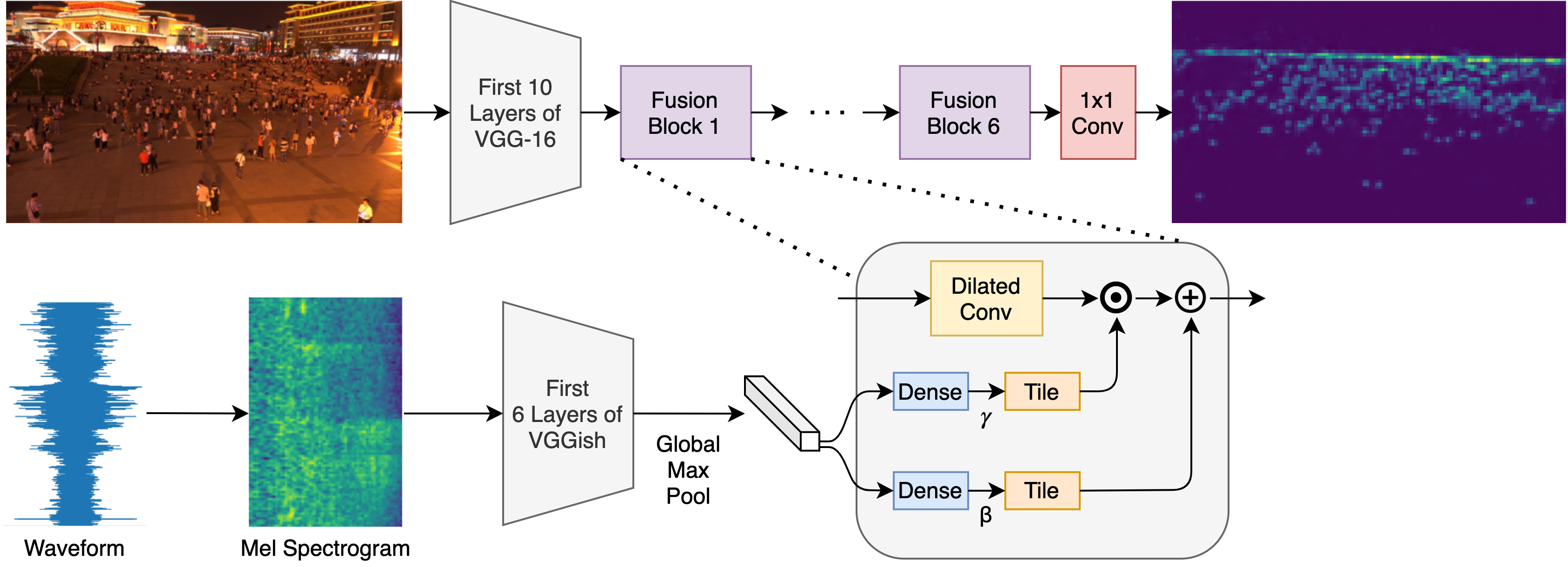}}
\caption{Architecture of AudioCSRNet.}
\label{audiocsrnet}
\end{figure*}

In CSRNet, the features extracted from the input image by the first 10 layers of a VGG-16 \cite{DBLP:journals/corr/SimonyanZ14a} network pre-trained on the ImageNet dataset \cite{Deng2009} are passed through 6 dilated convolution layers and a $ 1\times1 $ convolution layer in order to obtain the density map. AudioCSRNet extends this architecture by converting each of the dilated convolution layers into a fusion block. The architecture of AudioCSRNet is depicted in Figure \ref{audiocsrnet}. First, a Mel spectrogram is obtained from the raw audio waveform. Subsequently, in each fusion block, the features extracted from the input Mel spectrogram by the first 6 layers of a VGGish \cite{Hershey2017} network pre-trained on the AudioSet dataset \cite{Hershey2017} are projected to two vectors called $ \gamma $ and $ \beta $ which represent the multiplicative and additive aspects of the audio features. The $ \gamma $ and $ \beta $ vectors are then tiled in order to match the size of the visual features. Finally, the output of the dilated convolution is element-wise multiplied by $ \gamma $ and added to $ \beta $.

The fusion operation can be summarized as
\begin{equation}
v_{l + 1} = \mathcal{F}_l(\gamma_l \odot D_l(v_l) + \beta_l),
\label{audiocsrnet_fusion}
\end{equation}
where $ v_l \in \mathbb{R}^{C_l \times W_l \times H_l} $ is the output of the $ l $-th fusion block, $ F_l $ denotes an activation function, $ \gamma_l $ and $ \beta_l $ are the tiled vectors and $ D_l $ represents the $ l $-th dilated convolution.

In practice, a batch normalization layer \cite{pmlr-v37-ioffe15} is added immediately after each dilated convolution. Furthermore, the height and width of the intermediate features remain unchanged by using padding in the convolution operations, meaning $ H_l = H_{l + 1} $ and $ W_l = W_{l + 1} $. Additionally, since the first 10 layers of VGG-16 decrease both height and width by a factor of 8 via several pooling operations, the final result of the network needs to be upsampled by a factor of 8 in order to match the resolution of the input image. It is important to preserve the total sum of the density map during this upsampling operation, since it represents the total count.

\section{Single-Layer Vision Transformers for Early Exits}\label{S:Method}

We assume a pre-trained and high performing backbone network is already available. Due to time constraints arising from the particular application, it is desirable that the network provides a result within the specific deadline rather than not providing a result at all, even though this result may be less accurate than it would be if time constraints did not exist. Therefore, the backbone needs to be augmented with early exit branches to allow for dynamic inference and anytime prediction. As previously mentioned, we use the classifier-wise approach for training the early exit branches since it results in ``plug-and-play'' branches that can easily be added to the backbone network without any re-training or hyper-parameter tuning.

\subsection{SL-ViT}

\begin{figure}[htbp]
\centerline{\includegraphics[width=0.48\textwidth]{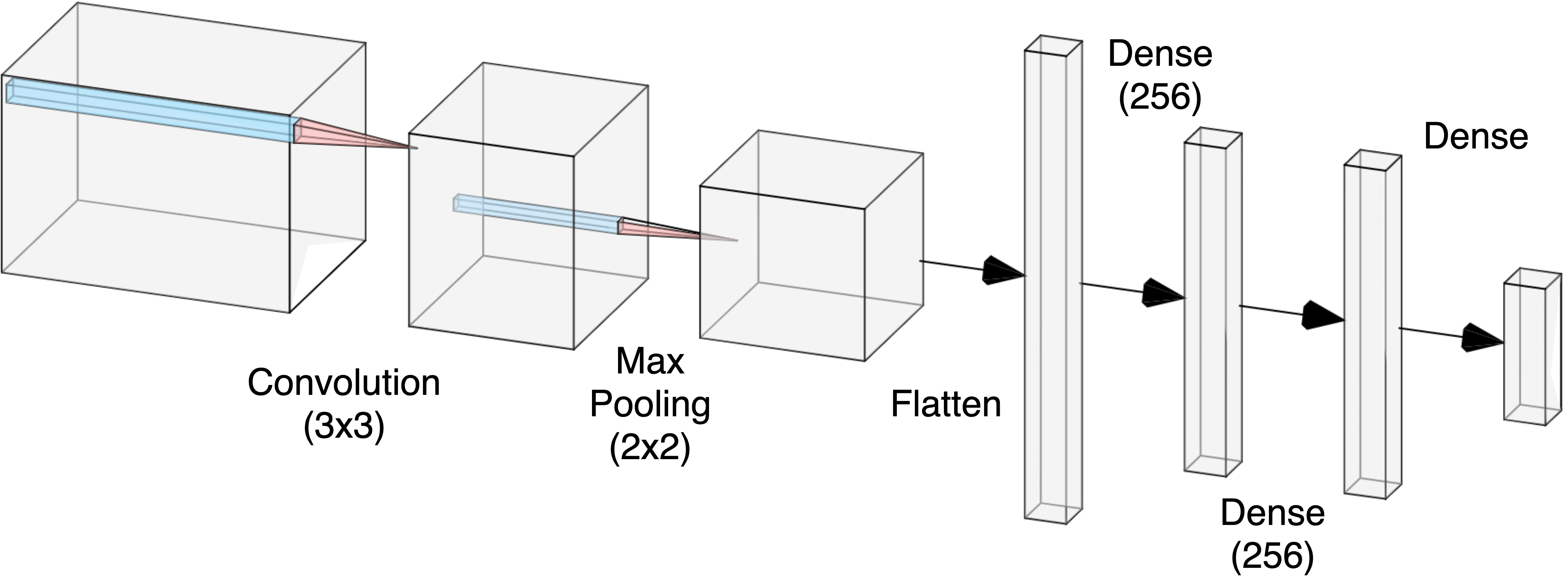}}
\caption{Architecture of CNN early exit branches. Size of the flattened feature vector depends on the dimensions of the features at the specific branch location. For branches placed on the AudioCSRNet backbone, max pooling size is increased to 4x4 since the input images have a high resolution. Figure created using the NN-SVG tool \cite{LeNail2019}.}
\label{cnn_branch_arch}
\end{figure}

Typically, the architecture of early exit branches starts with one or more convolution layers, although some may have no convolutions at all. Afterwards, they may have a pooling layer, which may be global pooling, and one MLP \cite{tkhu2019triplewins, Teerapittayanon2016}. Here, as a baseline, we choose to utilize the architecture depicted in Figure \ref{cnn_branch_arch} with one $ 3 \times 3 $ convolution, followed by a $ 2 \times 2 $ max pooling layer and finally a MLP. The size of the max pooling layer is increased to $ 4 \times 4 $ for crowd counting since the input images have a very high resolution. Additionally, we use dropout \cite{10.5555/2627435.2670313} inside the MLP to avoid overfitting. We use a single convolution since early exits with two or more convolution layers have a high overhead and may even lead to lower accuracy \cite{Teerapittayanon2016}. Early exits without convolutions are sometimes used very late in the network, however, since they are straightforward and leave no room for modifications, we do not apply our method for such cases. The resulting architecture is a common setup within the literature, and is effectively the same architecture used for earlier exits by Hu et al. \cite{tkhu2019triplewins}.

Our method called \textit{single-layer vision transformer} or \textit{SL-ViT} for short, is an alternative architecture for early exit branches that can achieve a higher accuracy compared to the aforementioned baseline, while having less overhead in terms of the number of parameters as well as floating point operations per second (FLOPS). Our proposed architecture is based on the vision transformer architecture introduced in section \ref{S:RelatedWork_ViT}, where instead of the input image, we split the intermediate features at the branch location into patches (sub-tensors) and pass them to a vision transformer.

The choice of vision transformer architecture is primarily due to its global receptive field. Receptive field is crucial in many deep learning problems, including ones studied in this work. The receptive field of state-of-the-art CNNs developed for image classification has steadily increased over time and is correlated with increased classification accuracy \cite{araujo2019computing}. Additionally, in audio classification using spectrograms, each location relates to a different frequency band in a different window of time. It is reasonable to assume that processing combinations of frequencies and windows that are not necessarily adjacent could be of importance. Moreover, many crowd counting methods have made use of global information through visual attention mechanisms and dilated convolutions \cite{2003.12783}. Since the receptive field is particularly limited in early layers of CNN backbones, choosing an architecture for early exit branches with a global receptive field could be beneficial.

Many other designs strive to increase the receptive field in their building blocks, for instance, the \textit{pyramid pooling module (PPM)} in PSPNet \cite{8100143} or \textit{atrous spatial pyramid pooling (ASPP)} in DeepLab \cite{7913730}. However, they all fall short in comparison with the global receptive field of transformers. PPM increases the receptive field through aggregating different levels of pooling, which means far locations have only access to coarse representations of each other, and ASPP has holes in its receptive field.

It is important to mention that the local receptive field of convolutional layers is not fundamentally bad. On the contrary, it plays a key role in representation learning and extracting local information, especially in the early layers of the network where the receptive field of the convolutional filters is small. Filters in successive convolutional layers have increasingly larger receptive fields, therefore, final layers in a CNN architecture have filters of large enough receptive fields that can effectively aggregate information from the entire input image to provide a proper response. However, this process of cascading local receptive fields of increasing size requires the number of layers in the CNN to be large, or at least all the layers in the network to be traversed in order to provide the network’s response. When an early exit is added at an early layer, this chain of increasingly larger receptive fields is broken, and an early exit that has a local receptive field may not be able to effectively aggregate all required information in the image to provide a suitable response. This situation is the motivation behind the proposed branch architecture, which fuses the local receptive field of the layer in the network where the early exit branch is attached, with the global receptive field of the early exit, in order to effectively aggregate information from the entire input and provide a more accurate response. Indeed, the original Vision Transformer paper \cite{dosovitskiy2021an} attributes the success of their model to the combination of local and global receptive fields and shows that even in very early layers, this ability to integrate information globally is indeed used by the model.

\begin{figure}[htbp]
\centerline{\includegraphics[width=0.48\textwidth]{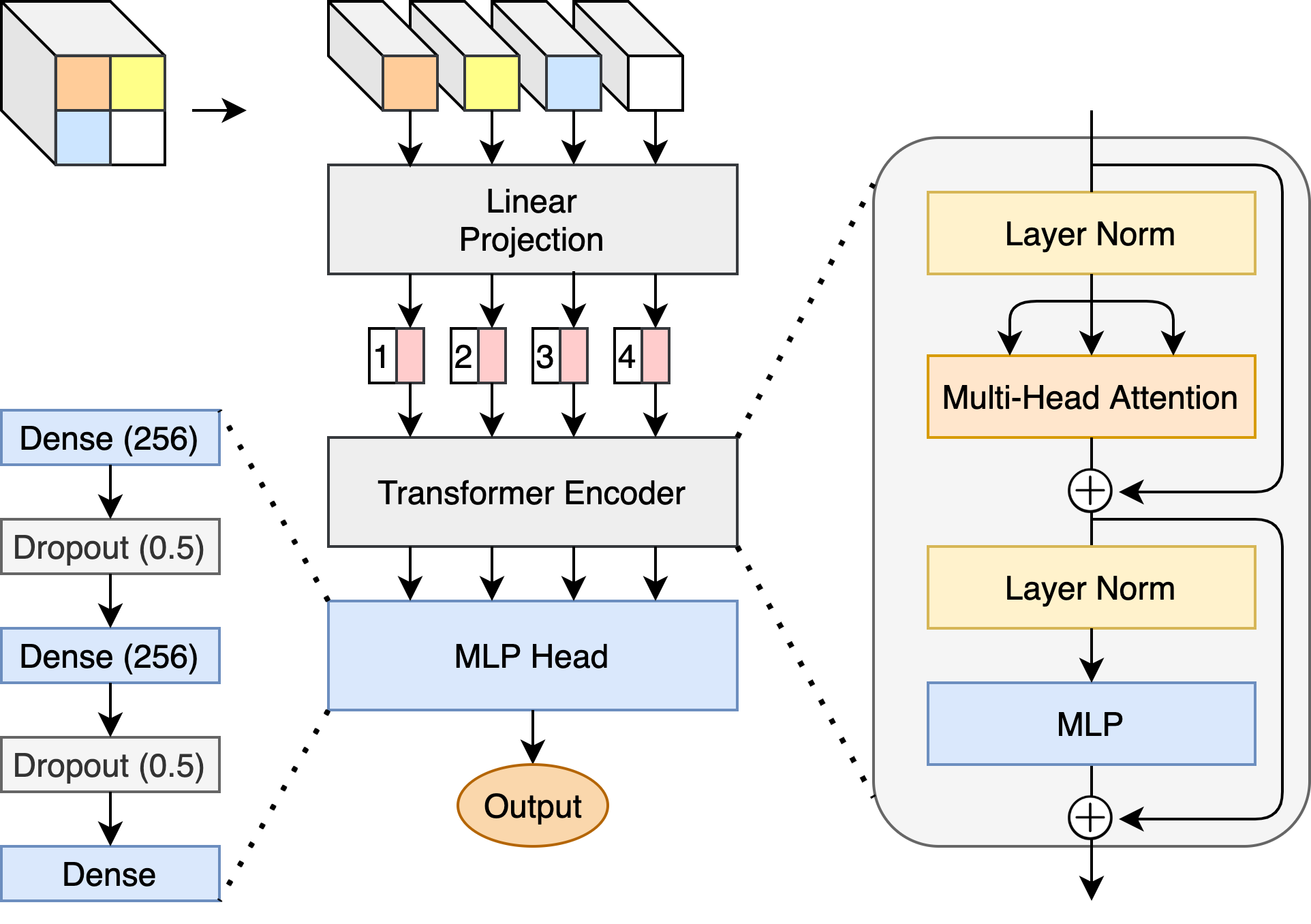}}
\caption{Architecture of SL-ViT early exit branches. Unlike typical vision transformers, only a single transformer encoder layer is used, extra learnable classification token is not added to the sequence and the entire output of the transformer encoder is passed on to the MLP head. The MLP head is the same as CNN early exit branches.}
\label{sl_vit_branch_arch}
\end{figure}

There are some crucial differences between the original vision transformer and the architecture in our method. First, in order to introduce a low overhead for early exit branches, we only use a single transformer encoder layer instead of the original 12 to 36 layers, meaning that $ L = 1 $ in our case. Secondly, we do not utilize a separate classification token and instead pass the entire output of the transformer encoder layer to the MLP head. This is possible because the width and height of tensors are generally reduced throughout CNN backbones by pooling operations, and thus the number of patches in our architecture is lower than that of the original vision transformer. In addition to the number of patches, the size of the embedding dimension ($ d $) is also reduced in our proposed architecture, introducing far less parameters when passing the entire output of the last transformer encoder layer to the MLP head, even with high-resolution inputs such as in our crowd counting experiments. Variations of our architecture have $ 5 \times 5 $, $ 7 \times 7 $ or $ 16 \times 9 $ patches and embedding dimensions of 32 or 36, whereas different versions of the original vision transformer have $ 14 \times 14 $ or $ 16 \times 16 $ patches and embedding dimensions of 768, 1024 or 1280. We empirically found that using the entire transformer encoder output instead of just one classification token can increase the accuracy, perhaps because in a single-layer version, there are not enough layers for the classification token to learn to properly summarize other patches. Our proposed architecture is shown in Figure \ref{sl_vit_branch_arch}. It is also important to note that the MLP head used in our architecture is exactly the same as the MLP in the CNN early exit architecture.

Our model has several hyper-parameters, namely the size of each patch, the embedding dimension $ d $ and the number of attention heads $ h $ in multi-head attention. The patch size creates a trade-off where smaller patches result in a more fine-grained attention mechanism while increasing the total number of parameters in a bi-quadratic fashion. Therefore, similar to the original vision transformer, we choose the size of the patch to be close to the square root of the height and width of the input features. We also make sure that the size of the patch can divide the size of the input features to avoid padding, for instance, a patch size of $ 4 \times 4 $ for input features of size $ 28 \times 28 $. We perform a grid search to find the values of $ d $ and $ h $ that result in the highest accuracy, while keeping the total number of parameters less than or equal to that of the CNN early exit counterpart.

At a first glance, it might seem like the SL-ViT architecture introduces more hyper-parameters than the conventional CNN architecture, however, the CNN architecture includes many design choices as well, such as the number of filters, filter size, padding, dilation, stride, pooling type and pooling size. The design choices for CNN architectures might seem simpler since they have been studied more extensively compared to vision transformers which were introduced more recently.

\subsection{Audiovisual SL-ViT}

With audiovisual backbones such as the AudioCSRNet model for audiovisual crowd counting, described in section \ref{S:RelatedWork_AudioCSRNet}, it is desirable to have audiovisual early exits that use both visual and audio features in order to achieve a higher accuracy. The simplest way to have such branches is to add the branches after the blocks where the fusion of visual and audio features take place. However, with our proposed SL-ViT architecture, it is also possible to include audio features as one or more patches alongside other patches, and directly fuse the features in the early exit. The advantage of this approach is that since in vision transformers, any of the patches can pay attention to any other patch, the visual features can be fused with the audio features without being directly impacted and modified. In contrast, since convolutional filters only take the immediate vicinity into account, the audio features must be present in every location. One option is to concatenate the visual features and the tiled audio features along the depth. However, that would greatly increase the amount of computation for each fusion operation, therefore intrusive operations such as element-wise multiplication and addition are used instead.

\subsection{Copycat SL-ViT}

Finally, we introduce a fine-tuning strategy for SL-ViT branches that can further increase their accuracy. Correia-Silva et al. \cite{CorreiaSilva2018} developed a method called \textit{copycat CNN} where they create a ``fake'' dataset by taking images from another domain, giving them as input to a network trained on the target domain, and recording the output of the network as labels for these images. For instance, images from the ImageNet dataset \cite{Deng2009} can be given to a network trained on the CIFAR-10 dataset \cite{Krizhevsky2009}, where the image of a camel may be labelled as a ``dog'' since there are no labels for ``camel'' in CIFAR-10. This fake dataset is then combined with a dataset for the target domain and used to train a new network. We use this strategy to fine-tune an already trained SL-ViT branch and obtain a \textit{copycat single-layer vision transformer} (\textit{CC-SL-ViT}). Note that the ratio of the fake data mixed with the available dataset is a hyper-parameter of this fine-tuning strategy.

\section{Experimental Setup}\label{S:ExperimentalSetup}
In this section, we provide the details of our experiments. We begin by giving a short summary of the datasets as well as the training details for the backbone networks. We then lay out the details of the branch architectures, their training procedure and their placement on the backbone networks, and finally explain how the copycat strategy was used to fine-tune the branches.

A total of 27 different scenarios were tested in our experiments. For both image and audio classification, two datasets, three backbone networks and two different branch locations on each backbone were tested. In addition, three different branch locations for the audiovisual crowd counting backbone network were covered. All experiments were repeated 5 times and the average accuracy as well as the standard deviation were recorded. 4 $ \times $ Nvidia 2080Ti GPUs were used for the training of our models.

\subsection{Datasets}
\subsubsection{CIFAR-10 and CIFAR-100}
These are widely-used datasets for image classification \cite{Krizhevsky2009}. Both datasets consist of 60,000 color images of size $ 32 \times 32 $ pixels and their corresponding class labels. The images in CIFAR-10 and CIFAR-100 are categorized into 10 and 100 different classes, respectively. We use 40,000 examples for training, 10,000 for validation and another 10,000 for testing. Since our backbone networks are pre-trained on ImageNet which consists of $ 224 \times 224 $ pixel images, we resize each image to these dimensions before passing them into the network.

\subsubsection{Speech Commands (SC)} A well-known audio dataset of spoken words \cite{1804.03209}. It consists of 100,503 1-second audio clips with a sampling rate of 16kHz, each labelled as one of 12 classes: 10 different spoken words such as ``Yes'', ``No'', ``Down'' and ``Stop'' as well as one class for background noise and another for unknown words. We use 85,511 examples for training, 10,102 for validation and 4,890 for testing. We convert the raw audio waveforms into spectrograms using short-time Fourier transform (STFT) with a window size of 255 samples and step size of 128 samples, and resize the resulting spectrograms to $ 224 \times 224 $ before passing them into the network.

\subsubsection{GTZAN} It is the most widely-used dataset for music genre recognition \cite{Tzanetakis2002}. The original dataset consists of 10 genres such as ``Pop'' and ``Rock'' and 100 30-second audio clips per genre with a sampling rate of 22,050Hz. We follow the common approach to split each audio clip into 10 separate 3-second clips in order to increase the size of the dataset to 10,000. We use 8,000 examples for training, 1,000 for validation and another 1,000 for testing. Following the approach of Palanisamy et al. \cite{2007.11154} where different spectrograms with different parameters are placed in each channel of the input image, we use one spectrogram obtained from STFT with window size of 512 samples and step size of 256 samples as well as two Mel spectrograms with 128 bins and window sizes of 100 and 50 milliseconds, and step sizes of 50 and 25 milliseconds, respectively.

\subsubsection{DISCO} An audiovisual dataset for crowd counting which contains 1,935 images of Full HD resolution ($ 1920 \times 1080 $) \cite{2005.07097}. For each image, a corresponding 1-second audio clip of ambient sounds with a sampling rate of 48kHz, starting 0.5 seconds before the image was taken and ending 0.5 seconds afterwards, exists as well. The labels are provided in the form of head annotations in the image. At the time of this writing, DISCO is the only publicly available dataset for audiovisual crowd counting. We use 1435 examples for training, 200 for validation and 300 for testing. The input image is resized to $ 1024 \times 576 $ pixels to reduce memory and computation requirements. Similar to Hershey et al. \cite{Hershey2017}, the input audio waveform is transformed into a Mel spectrogram with 64 bins, window size of 25 milliseconds and step size of 10 milliseconds. Following Hu et al. \cite{2005.07097} the ground truth density maps are obtained by convolving the head annotations with a $ 15 \times 15 $ Gaussian kernel $ \mathcal{K} \sim \mathcal{N}(0, 4.0) $.

\subsection{Backbone networks}
Transfer learning is used to train the ResNet152, \\DenseNet201 and InceptionV3 backbone networks for both image and audio classification. The backbone networks are all pre-trained on the ImageNet dataset and the top layer is replaced. We found that instead of adding just one dense layer at the top, as is common in transfer learning, using two dense layers and a dropout layer in between leads to a higher accuracy in our case. The resulting network is then trained using the Adam optimizer \cite{DBLP:journals/corr/KingmaB14} with a learning rate of $ 10^{-4} $ and categorical cross-entropy loss function. The learning rate is reduced by a factor of 0.6 on plateau with a tolerance of 2 epochs, and an early stopping mechanism with a tolerance of 5 epochs is used.

The audiovisual crowd counting backbone is trained in two stages. We first train a network with the AudioCSRNet architecture described in Section \ref{S:RelatedWork_AudioCSRNet} for 100 epochs. $ L_2 $ norm is used as loss function and AdamW \cite{DBLP:conf/iclr/LoshchilovH19} with a learning rate of $ 10^{-5} $ and weight decay of $ 10^{-4} $ is used as optimizer, where the learning rate is multiplied by a factor of 0.99 each epoch. This is the same training procedure used in the original paper \cite{2005.07097}. Subsequently, in order to convert the problem from dense prediction to regression, a dense layer with an output size of one is added after the last layer of the trained AudioCSRNet. This layer is initialized as a sum, meaning the initial weights are all equal to one and no bias is used. Then the entire network is re-trained for another 100 epochs using MAE as loss function instead of the previous $ L_2 $ loss, a learning rate of $ 10^{-6} $ and weight decay of $ 10^{-5} $. The learning rate is similarly multiplied by a factor of 0.99 every epoch. The resulting model achieves a MAE of 13.63 which is even lower than the MAE of 14.27 reported in the original paper. However, the output of the network is just a single number representing the total count instead of a density map. The final accuracy of all trained backbones can be seen in Table \ref{backbone_performance}. 

When training the backbone networks, in order to fit the limitations of our available computational resources, the batch sizes are adjusted and some layers of the backbone networks are frozen. All backbone networks were trained with a batch size of 32 except AudioCSRNet which has a batch size of 4 as well as InceptionV3 when trained on CIFAR-10 and CIFAR-100 which has a batch size of 64. All layers of the backbone networks were trained, except in the case of ResNet152 and DenseNet201 when trained on CIFAR-10 and CIFAR-100 where only the batch normalization layers were trained. We found that training only the batch normalization layers is sufficient to achieve a high-performing backbone network in these cases \cite{2003.00152}.

\begin{table}[htbp]
\caption{Performance of backbone networks on each dataset}
\begin{center}
\resizebox{\linewidth}{!}{
\begin{tabular}{ c | c c c c c } 
\hline
    Backbone &
    CIFAR-10 Acc. &
    CIFAR-100 Acc. &
    SC Acc. &
    GTZAN Acc. &
    DISCO MAE\\
\hline
\hline
    ResNet152 &
    95.36\% &
    82.25\% &
    95.85\% &
    91.29\% &
    -\\
    
    DenseNet201 &
    96.48\% &
    82.53\% &
    96.36\% &
    92.09\% &
    -\\
    
    InceptionV3 &
    96.56\% &
    83.80\% &
    94.93\% &
    87.79\% &
    -\\
    
    AudioCSRNet &
    - &
    - &
    - &
    - &
    13.63\\

\hline
\end{tabular}
}
\end{center}
\label{backbone_performance}
\end{table}

\subsection{Branches}

All branches were trained from scratch using the He initialization method \cite{he2015delving} and the Adam optimizer with a learning rate of $ 10^{-4} $ where the learning rate is reduced by a factor of 0.6 on plateau with a tolerance of 2 epochs, and an early stopping mechanism with a tolerance of 5 epochs is utilized. The branches on classification backbones use a categorical cross-entropy loss function whereas the branches on the audiovisual crowd counting backbone use mean absolute error loss. The training batch size for branches were 64 in scenarios involving CIFAR-10, CIFAR-100 and Speech Commands, 32 in scenarios involving \\GTZAN and 4 in scenarios involving DISCO.

Table \ref{exit_locations} shows the location of the branches placed on each backbone network. For the AudioCSRNet backbone network, branch V1 uses only the output of the VGG-16 layers, therefore, it only has access to the visual features. Branch AV1 uses the outputs of both VGG-16 and VGGish, therefore it has access to both audio and visual features. In this branch location, the fusion of audio and visual features is performed as described in Section \ref{S:Method} for the SL-ViT architecture, and similar to the fusion blocks in AudioCSRNet for the CNN architecture, however, without dilation. Finally, branch AV2 is placed after the first fusion block in AudioCSRNet, therefore audio and visual features have already been fused and thus fusion operation is not required within the branches. Adding branches after the second fusion block or later would not be reasonable since more than 85\% of the computation of the backbone is carried out before that point, and thus the acceleration resulting from early exits would be negligible.

\begin{table}[htbp]
\caption{Placement of branches for each backbone betwork}
\begin{center}
\resizebox{\linewidth}{!}{
\begin{tabular}{ c c | c } 
\hline
    Backbone &
    BN$^{*}$ &
    Branch Placed After \\
\hline
\hline
    DenseNet201 &
    1 &
    Transition Layer 1 \\
    
    &
    2 &
    Transition Layer 2 \\

\hline

    ResNet152 &
    1 &
    12th Convolution \\

    &
    2 &
    36th Convolution \\

\hline

    InceptionV3 &
    1 &
    First Filter Concat \\

    &
    2 &
    Second Filter Concat \\

\hline

    AudioCSRNet &
    V1 &
    Last Layer of VGG \\

    &
    AV1 &
    Last Layers of VGG and VGGish\\

    &
    AV2 &
    First Fusion Block \\

\hline
\multicolumn{3}{l}{$^{*}$Branch Number} \\
\end{tabular}
}
\end{center}
\label{exit_locations}
\end{table}

\subsection{SL-ViT and CC-SL-ViT Parameters}
Table \ref{sl_vit_parameters} summarizes the hyper-parameters used for the SL-ViT branches in each scenario. ``Patch Size'' shows the width and height of each image patch, ``Patches'' denotes the resulting number of patches across width and height of the input image, $ d $ is the size of embedding dimension and $ h $ is the number of heads in multi-head attention.

For copycat SL-ViT, images from the Tiny ImageNet dataset, which are the images from ImageNet down-sampled to $ 32 \times 32 $, were given to the InceptionV3 backbone trained on CIFAR-10, and the outputs were used to create the fake dataset. Then the fake dataset was mixed with CIFAR-10 with a 2-to-1 ratio and used for re-training.

\begin{table}[htbp]
\caption{Hyper-parameters of SL-ViT for different backbone networks and branches}
\begin{center}
\resizebox{\linewidth}{!}{
\begin{tabular}{ c c c | c c c c } 
\hline
    Backbone &
    Dataset &
    BN$^{*}$ &
    Patch Size &
    Patches &
    $ d $ &
    $ h $ \\
\hline
\hline
    DenseNet201 &
    all &
    all &
    4x4 &
    7x7 &
    32 &
    12 \\

\hline

    ResNet152 &
    SC &
    2 &
    4x4 &
    7x7 &
    32 &
    24 \\

    &
    GTZAN &
    2 &
    4x4 &
    7x7 &
    32 &
    24 \\

    &
    \multicolumn{2}{c|}{Other} &
    4x4 &
    7x7 &
    32 &
    12 \\

\hline

    InceptionV3 &
    CIFAR-100 &
    all &
    5x5 &
    5x5 &
    36 &
    8 \\

    &
    \multicolumn{2}{c|}{Other} &
    5x5 &
    5x5 &
    32 &
    12 \\

\hline

    AudioCSRNet &
    DISCO &
    all &
    8x8 &
    16x9 &
    32 &
    12 \\

\hline
\multicolumn{3}{l}{$^{*}$Branch Number} \\
\end{tabular}
}
\end{center}
\label{sl_vit_parameters}
\end{table}

\section{Results}\label{S:Results}
The results of our experiments are presented in Tables \ref{cifar10_results} to \ref{disco_results}. In these Tables, the final accuracy, the total FLOPS of the model up to and including the branch and the number of parameters of just the early exit branch are compared between the CNN architecture and the SL-ViT architecture. Higher accuracies, lower errors, lower number of parameters and lower total FLOPS are highlighted in these tables. Furthermore, the acceleration caused by SL-ViT early exits, defined as the total FLOPS of the backbone network divided by the total FLOPS of the model up to and including the SL-ViT branch, is also provided.

Several observations can be made about these results. First, in all scenarios except one, SL-ViT early exits achieve a significantly higher accuracy. Even in the one exceptional scenario, namely branch 2 of ResNet152 in Table \ref{speech_commands_results}, the accuracy of SL-ViT is very close to its CNN counterpart. Secondly, while in some cases SL-ViT branches have an equal number of parameters compared to CNN branches, in all scenarios, the total FLOPS of SL-ViT branches is lower, therefore SL-ViT branches are always more lightweight. Thirdly, in one scenario, namely branch 2 of ResNet152 in Table \ref{gtzan_results}, removing the last residual connection in the SL-ViT architecture significantly improved the accuracy of the SL-ViT branch. Finally, in the AV2 branch location in Table \ref{disco_results}, both CNN and SL-ViT are impractical branches since earlier branches with higher accuracies exist. This is perhaps due to the intrusive fusion operation in the first fusion block which might initially make the intermediate features more obscure. Nonetheless, even in this case, SL-ViT is more accurate.

\begin{table*}[htbp]
\caption{Comparison of different early exit architectures on the CIFAR-10 dataset}
\begin{center}
\resizebox{\linewidth}{!}{
\begin{tabular}{ c c | c c | c c | c c | c } 
\hline
    Backbone &
    Branch &
    \multicolumn{2}{c}{Accuracy} &
    \multicolumn{2}{c}{Branch Params} &
    \multicolumn{2}{c}{Total FLOPS} &
    Acceleration \\
    &
    &
    CNN &
    SL-ViT &
    CNN &
    SL-ViT &
    CNN &
    SL-ViT & 
    SL-ViT\\
\hline
\hline
    ResNet152 &
    1 &
    66.74 ± 0.57\% &
    \textbf{70.79 ± 0.72\%} &
    0.78M &
    \textbf{0.59M} &
    1.66B &
    \textbf{1.64B} &
    13.77 \\
    
    &
    2 &
    79.31 ± 0.81\% &
    \textbf{81.18 ± 0.52\%} &
    0.83M &
    \textbf{0.79M} &
    5.33B &
    \textbf{5.26B} &
    4.29 \\

    DenseNet201 &
    1 &
    71.27 ± 0.36\% &
    \textbf{76.38 ± 0.33\%} &
    0.78M &
    \textbf{0.59M} &
    2.55B &
    \textbf{2.53B} &
    3.39 \\
    
    &
    2 &
    80.64 ± 0.29\% &
    \textbf{83.53 ± 0.37\%} &
    0.80M &
    \textbf{0.66M} &
    4.21B &
    \textbf{4.17B} &
    2.06 \\

    InceptionV3 &
    1 &
    77.27 ± 0.58\% &
    \textbf{79.99 ± 0.20\%} &
    0.61M &
    \textbf{0.56M} &
    2.17B &
    \textbf{2.14B} &
    2.65 \\
    
    &
    2 &
    79.55 ± 0.24\% &
    \textbf{81.72 ± 0.53\%} &
    0.61M &
    \textbf{0.56M} &
    2.53B &
    \textbf{2.49B} &
    2.28 \\
    
\hline
\end{tabular}
}
\end{center}
\label{cifar10_results}
\end{table*}

\begin{table*}[htbp]
\caption{Comparison of different early exit architectures on the CIFAR-100 dataset}
\begin{center}
\resizebox{\linewidth}{!}{
\begin{tabular}{ c c | c c | c c | c c | c } 
\hline
    Backbone &
    Branch &
    \multicolumn{2}{c}{Accuracy} &
    \multicolumn{2}{c}{Branch Params} &
    \multicolumn{2}{c}{Total FLOPS} &
    Acceleration \\
    &
    &
    CNN &
    SL-ViT &
    CNN &
    SL-ViT &
    CNN &
    SL-ViT &
    SL-ViT \\
\hline
\hline
    ResNet152 &
    1 &
    34.93 ± 0.52\% &
    \textbf{38.59 ± 1.40\%} &
    0.80M &
    \textbf{0.61M} &
    1.66B &
    \textbf{1.64B} &
    13.77 \\
    
    &
    2 &
    47.39 ± 0.65\% &
    \textbf{53.93 ± 0.68\%} &
    0.86M &
    \textbf{0.81M} &
    5.33B &
    \textbf{5.26B} &
    4.29 \\
    
    DenseNet201 &
    1 &
    33.91 ± 1.00\% &
    \textbf{42.50 ± 0.69\%} &
    0.80M &
    \textbf{0.61M} &
    2.55B &
    \textbf{2.53B} &
    3.39 \\
    
    &
    2 &
    47.22 ± 0.45\% &
    \textbf{50.76 ± 1.01\%} &
    0.82M &
    \textbf{0.68M} &
    4.21B &
    \textbf{4.17B} &
    2.06 \\
    
    InceptionV3 &
    1 &
    43.18 ± 0.69\% &
    \textbf{46.86 ± 0.57\%} &
    0.63M &
    0.63M &
    2.17B &
    \textbf{2.14B} &
    2.66 \\
    
    &
    2 &
    44.87 ± 0.83\% &
    \textbf{49.07 ± 0.55\%} &
    0.63M &
    0.63M &
    2.53B &
    \textbf{2.50B} &
    2.28 \\
    
\hline
\end{tabular}
}
\end{center}
\label{cifar100_results}
\end{table*}

\begin{table*}[htbp]
\caption{Comparison of different early exit architectures on the Speech Commands dataset}
\begin{center}
\resizebox{\linewidth}{!}{
\begin{tabular}{ c c | c c | c c | c c | c } 
\hline
    Backbone &
    Branch &
    \multicolumn{2}{c}{Accuracy} &
    \multicolumn{2}{c}{Branch Params} &
    \multicolumn{2}{c}{Total FLOPS} &
    Acceleration \\
    &
    &
    CNN &
    SL-ViT &
    CNN &
    SL-ViT &
    CNN &
    SL-ViT &
    SL-ViT \\
\hline
\hline
    ResNet152 &
    1 &
    75.80 ± 0.73\% &
    \textbf{84.05 ± 0.31\%} &
    0.78M &
    \textbf{0.59M} &
    1.66B &
    \textbf{1.64B} &
    13.77 \\
    
    &
    2 &
    \textbf{89.78 ± 0.24\%} &
    89.63 ± 0.52\% &
    0.84M &
    0.84M &
    5.33B &
    \textbf{5.26B} &
    4.29 \\
    
    DenseNet201 &
    1 &
    72.78 ± 0.64\% &
    \textbf{87.94 ± 0.85\%} &
    0.78M &
    \textbf{0.59M} &
    2.55B &
    \textbf{2.53B} &
    3.39 \\
    
    &
    2 &
    86.56 ± 0.61\% &
    \textbf{90.93 ± 0.52\%} &
    0.80M &
    \textbf{0.66M} &
    4.21B &
    \textbf{4.17B} &
    2.06 \\
    
    InceptionV3 &
    1 &
    84.64 ± 0.88\% &
    \textbf{87.62 ± 0.65\%} &
    0.61M &
    \textbf{0.56M} &
    2.17B &
    \textbf{2.14B} &
    2.65 \\
    
    &
    2 &
    87.08 ± 1.11\% &
    \textbf{88.33 ± 0.92\%} &
    0.61M &
    \textbf{0.56M} &
    2.53B &
    \textbf{2.49B} &
    2.28 \\
    
\hline
\end{tabular}
}
\end{center}
\label{speech_commands_results}
\end{table*}

\begin{table*}[htbp]
\caption{Comparison of different early exit architectures on the GTZAN dataset}
\begin{center}
\resizebox{\linewidth}{!}{
\begin{tabular}{ c c | c c | c c | c c | c} 
\hline
    Backbone &
    Branch &
    \multicolumn{2}{c}{Accuracy} &
    \multicolumn{2}{c}{Branch Params} &
    \multicolumn{2}{c}{Total FLOPS} &
    Acceleration \\
    &
    &
    CNN &
    SL-ViT &
    CNN &
    SL-ViT &
    CNN &
    SL-ViT &
    SL-ViT \\
\hline
\hline
    ResNet152 &
    1 &
    67.01 ± 1.11\% &
    \textbf{73.27 ± 0.91\%} &
    0.78M &
    \textbf{0.59M} &
    1.66B &
    \textbf{1.64B} &
    13.77 \\
    
    &
    2$^{*}$ &
    80.26 ± 2.07\% &
    \textbf{81.56 ± 1.57\%} &
    0.83M &
    0.83M &
    5.33B &
    \textbf{5.26B} &
    4.29 \\
    
    DenseNet201 &
    1 &
    70.65 ± 1.23\% &
    \textbf{76.38 ± 1.94\%} &
    0.78M &
    \textbf{0.59M} &
    2.55B &
    \textbf{2.53B} &
    3.39 \\
    
    &
    2 &
    81.72 ± 0.62\% &
    \textbf{84.00 ± 1.67\%} &
    0.80M &
    \textbf{0.66M} &
    4.21B &
    \textbf{4.17B} &
    2.06 \\
    
    InceptionV3 &
    1 &
    77.86 ± 0.90\% &
    \textbf{79.42 ± 0.99\%} &
    0.61M &
    \textbf{0.56M} &
    2.17B &
    \textbf{2.14B} &
    2.65 \\
    
    &
    2 &
    78.90 ± 0.90\% &
    \textbf{79.90 ± 0.79\%} &
    0.61M &
    \textbf{0.56M} &
    2.53B &
    \textbf{2.49B} &
    2.28 \\
    
\hline
\multicolumn{8}{l}{$^{*}$The last residual connection in the SL-ViT architecture was removed in this case}
\end{tabular}
}
\end{center}
\label{gtzan_results}
\end{table*}

\begin{table*}[htbp]
\caption{Comparison of Different Early Exit Architectures on the DISCO Dataset}
\begin{center}
\resizebox{\linewidth}{!}{
\begin{tabular}{ c c | c c | c c | c c | c } 
\hline
    Backbone &
    Branch &
    \multicolumn{2}{c}{MAE} &
    \multicolumn{2}{c}{Branch Params} &
    \multicolumn{2}{c}{Total FLOPS} &
    Acceleration \\
    &
    &
    CNN &
    SL-ViT &
    CNN &
    SL-ViT &
    CNN &
    SL-ViT &
    SL-ViT \\
\hline
\hline
    AudioCSRNet &
    V1 &
    16.99 ± 0.28 &
    \textbf{15.04 ± 0.71} &
    2.50M &
    \textbf{2.35M} &
    329.77B &
    \textbf{328.72B} &
    1.49 \\

    &
    AV1&
    17.00 ± 0.23 &
    \textbf{14.58 ± 0.64} &
    2.52M &
    \textbf{2.36M} &
    331.37B &
    \textbf{330.31B} &
    1.48 \\

    &
    AV2 &
    17.90 ± 0.25 &
    \textbf{17.03 ± 1.04} &
    2.50M &
    \textbf{2.35M} &
    374.86B &
    \textbf{373.81B} &
    1.31 \\

\hline

\end{tabular}
}
\end{center}
\label{disco_results}
\end{table*}

Table \ref{copycat_results} shows the result of applying the copycat fine-tuning strategy to SL-ViT branches for the CIFAR-10 dataset. Observe than in all cases, the accuracy is significantly increased compared to SL-ViT, which itself was more accurate than CNN based on Table \ref{cifar10_results}. We also tested this strategy for the CIFAR-100 dataset with 10-to-1, 2-to-1 and 1-to-1 ratios of fake and real data, however, neither improved the overall accuracy. Perhaps another mixing ratio, choice of dataset and network to generate the fake dataset, optimizer or hyper-parameters such as learning rate may result in improvements for CIFAR-100.

\begin{table}[htbp]
\caption{Effect of Copycat strategy demonstrated on the CIFAR-10 dataset}
\begin{center}
\resizebox{\linewidth}{!}{
\begin{tabular}{ c c | c c } 
\hline
    Backbone &
    Branch &
    \multicolumn{2}{c}{Accuracy} \\
    &
    &
    SL-ViT &
    CC-SL-ViT \\
\hline
\hline
    ResNet152 &
    1 &
    70.79 ± 0.72\% &
    \textbf{71.61 ± 0.45\%} \\

    &
    2 &
    81.18 ± 0.52\% &
    \textbf{83.41 ± 0.15\%} \\
    
    DenseNet201 &
    1 &
    76.38 ± 0.33\% &
    \textbf{78.34 ± 0.31\%} \\
    
    &
    2 &
    83.53 ± 0.37\% &
    \textbf{84.89 ± 0.43\%} \\
    
    InceptionV3 &
    1 &
    79.99 ± 0.20\% &
    \textbf{80.78 ± 0.23\%} \\
    
    &
    2 &
    81.72 ± 0.53\% &
    \textbf{82.20 ± 0.40\%} \\
    
\hline
\end{tabular}
}
\end{center}
\label{copycat_results}
\end{table}

\begin{table}[htbp]
\caption{Comparison of improvements gained by SL-ViT with gains from knowledge distillation for the CIFAR-10 dataset.}
\begin{center}
\resizebox{\linewidth}{!}{
\begin{tabular}{ c c | c c c } 
\hline
Backbone & Branch & CNN (Baseline) & CNN with KD & SL-ViT (Ours)\\
\hline
\hline
ResNet152 & 1 & 66.74 ± 0.57\% & 69.31 ± 0.28\% & \textbf{70.79 ± 0.72\%}\\
& 2 & 79.31 ± 0.81\% & 78.79 ± 0.61\% & \textbf{81.18 ± 0.52\%}\\
DenseNet201 & 1 & 71.27 ± 0.36\% & 73.93 ± 0.15\% & \textbf{76.38 ± 0.33\%}\\
& 2 & 80.64 ± 0.29\% & 81.56 ± 0.12\% & \textbf{83.53 ± 0.37\%}\\
InceptionV3 & 1 & 77.27 ± 0.58\% & 78.37 ± 0.34\% & \textbf{79.99 ± 0.20\%}\\
& 2 & 79.55 ± 0.24\% & 80.41 ± 0.43\% & \textbf{81.72 ± 0.53\%}\\
\hline
\end{tabular}
}
\end{center}
\label{kd_compare_results}
\end{table}

Even though other early exit methods focus on improving the training procedure and can be used in combination with our proposed architecture, comparing the improvements gained by utilizing such methods with improvements gained from our approach can still provide insights into the significance of architecture design for early exits. Table \ref{kd_compare_results} contains comparisons with knowledge distillation-based training similar to the method in. \cite{Phuong2019} for the CIFAR-10 dataset. Observe that in all cases, SL-ViT obtains a significantly higher accuracy compared to knowledge distillation.

\subsection{Ablation Studies}

Table \ref{ablation_results} showcases the effect of using different architectural parameters on the accuracy of both SL-ViT and CNN branches. Where not specified, the CNN early exits have a $ 3 \times 3 $ kernel size with no dilation, and the SL-ViT early exits have 12 attention heads, which are the baselines presented in previous tables. Other parameters such as the number of convolutional filters and padding size are adjusted accordingly in order to keep the number of parameters close to the baselines.

These results support our hypothesis that the improvements of SL-ViT are due to the fusion of local and global receptive fields. First, by increasing the number of attention heads in SL-ViT, the accuracy increases significantly while the parameters only slightly increase, hinting that learning multiple types of attention plays a major role in SL-ViT. Secondly, by increasing the CNN kernel size from $ 3 \times 3 $ to $ 15 \times 15 $ the accuracy is improved, yet it is still lower than that of SL-ViT. This is because even a large filter size does not provide a global receptive field. On the other hand, adding dilation to CNN decreases its accuracy compared to the CNN baseline. This is because dilated convolutions create holes in the receptive field, which increases the receptive field yet loses important local information. Thirdly, using two CNN layers also improves the accuracy compared to the CNN baseline, however, a higher gain in accuracy was achieved using a larger kernel size. Moreover, two SL-ViT layers still obtain a higher accuracy compared to two CNN layers while having a lower overhead in terms of parameters. Finally, we show that even if the backbone is not pre-trained on ImageNet and is trained completely from scratch, SL-ViT still obtains a higher accuracy compared to CNN.

\begin{table}[htbp]
\caption{Ablation studies: the effect of the number of attention heads, number of layers, dilation, kernel size and backbone pre-training on the accuracy of early exits placed on the first branch location of a ResNet152 backbone trained on the CIFAR-10 dataset}
\begin{center}
\resizebox{\linewidth}{!}{
\begin{tabular}{ c | c c c  } 
\hline
    Architecture Params &
    Accuracy &
    Branch Params &
    FLOPS \\
\hline
\hline
    SL-ViT (1 head) & 67.92 ± 0.86\% & 0.55M & 1.64B\\
    SL-ViT (2 heads) & 68.65 ± 0.90\% & 0.55M & 1.64B\\
    SL-ViT (4 heads) & 69.08 ± 1.07\% & 0.56M & 1.64B\\
    SL-ViT (8 heads) & 69.85 ± 1.12\% & 0.58M & 1.64B\\
    SL-ViT (12 heads) & \textbf{70.79 ± 0.72\%} & 0.59M & 1.64B\\
    SL-ViT (16 heads) & 70.76 ± 0.40\% & 0.61M & 1.64B\\
    CNN ($ 3 \times 3 $ kernel) & 66.74 ± 0.57\% & 0.78M & 1.66B\\
    CNN ($ 11 \times 11 $ kernel) & 69.71 ± 1.06 \%& 0.78M & 1.88B\\
    CNN ($ 15 \times 15 $ kernel) & 69.90 ± 0.68\% & 0.79M & 2.02B\\
    CNN (dilation 2) & 66.61 ± 0.47\% & 0.78M & 1.66B\\
    CNN (dilation 3) & 65.43 ± 0.32\% & 0.78M & 1.66B\\
\hline
    SL-ViT (2 layers) & \textbf{71.89 ± 0.75\%} & 0.65M & 1.64B\\
    CNN (2 layers) & 67.68 ± 1.06\% & 0.78M & 1.66B\\
\hline
    SL-ViT (no backbone pre-training) & \textbf{63.14 ± 0.57\%} & 0.59M & 1.64B\\
    CNN (no backbone pre-training) &  62.86 ± 0.99\% & 0.78M & 1.66B\\
\hline
\end{tabular}
}
\end{center}
\label{ablation_results}
\end{table}

Finally, we discovered that removing the second residual connection in the transformer encoder may lead to an increase in the overall accuracy of our method. This effect was moderate in most cases, yet quite significant in others. An example of this effect is shown in Table \ref{ablation_residual_results} for the Speech Commands dataset. We chose to keep the residual connection whenever the effect was moderate and only remove it if it leads to a significantly higher accuracy. Such cases are highlighted in our experiments (Table \ref{gtzan_results}).

\begin{table}[htbp]
\caption{Ablation studies: the effect of removing the last residual connection in the transformer encoder for the Speech Commands dataset}
\begin{center}
\resizebox{\linewidth}{!}{
\begin{tabular}{ c c | c c  } 
\hline
    Backbone &
    Branch Number &
    Accuracy of SL-ViT &
    Accuracy of SL-ViT without the Last Residual \\
\hline
\hline
    ResNet152 & 1& \textbf{84.05 ± 0.31\%} & 83.67 ± 0.85\%\\
    & 2 & \textbf{89.63 ± 0.52\%} & 85.79 ± 0.58\%\\
    DenseNet201 & 1 & 87.94 ± 0.85\% & \textbf{88.35 ± 0.24\%}\\
    & 2 & 90.93 ± 0.52\% & \textbf{91.08 ± 0.52\%}\\
    InceptionV3 & 1 & \textbf{87.62 ± 0.65\%} & 86.10 ± 0.32\%\\
    & 2 & \textbf{88.33 ± 0.92\%} & 88.21 ± 0.45\%\\
\hline
\end{tabular}
}
\end{center}
\label{ablation_residual_results}
\end{table}

\subsection{Early Exit Procedure}

Since our method improves the accuracy in all early exit locations, it provides improvements regardless of which early exit procedure is used. For instance, suppose a \\confidence-based method is used where the result of an early exit branch is selected as the final answer if it is confident enough. In this setting, our method will lead to faster inference on average, since more accurate branches lead to higher confidence.

Another example would be the anytime prediction setting explained in the introduction, for instance, an edge server which receives inputs from many IoT devices and needs to provide a response for each input within a strict deadline. The transmission time from the IoT devices to the server changes over time due to network congestion. Moreover, the computational workload of the server varies over time, therefore, the time budget available for each input is not known beforehand, and the inference can be interrupted at any moment. In this case, the output of the latest exit is used as the final answer. In such a setting, our method will lead to more accurate results and faster inference, since SL-ViT exits are more accurate and have less overhead.

To make this more clear, we have conducted experiments within the anytime prediction setting, where a random time budget is assigned to each image in the CIFAR-10 test set. We use the DenseNet backbone and the two branch locations specified in Table \ref{exit_locations}. We compare the average accuracy and FLOPS between the case where SL-ViT branches are used and the case where CNN branches are utilized. The results of these experiments are shown in Table \ref{early_exit_procedure}. It can be observed that the multi-exit network with SL-ViT branches achieves a significantly higher average accuracy while having lower average FLOPS.

\begin{table}[htbp]
\caption{Comparison of the average accuracy and FLOPS in the anytime prediction setting between a multi-exit DenseNet with SL-ViT early exits and one with CNN early exits.}
\begin{center}
\resizebox{\linewidth}{!}{
\begin{tabular}{ c | c c  } 
\hline
    Model &
    Average Accuracy &
    Average FLOPS \\
\hline
\hline
    Multi-Exit DenseNet with CNN Branches & 82.79 ± 0.17\% & 5.11B\\
    Multi-Exit DenseNet with SL-ViT Branches & \textbf{85.65 ± 0.21\%} & \textbf{5.09B}\\
\hline
\end{tabular}
}
\end{center}
\label{early_exit_procedure}
\end{table}

\section{Discussion and Conclusion}\label{S:Conclusions}
We showed that the proposed architecture for early exit branches, namely single-layer vision transformer (SL-ViT) can consistently obtain a significantly higher accuracy compared to conventional methods while introducing a lower overhead in terms of FLOPS. We showed that our method works for both classification and regression problems, in both single and multi-modal scenarios, and across different backbone networks and branch locations.

As previously mentioned, one possible explanation for why SL-ViT performs better, is the fact that even a single layer of transformer encoder has a global receptive field since each patch can attend to any other patch, while a convolutional layer has a limited receptive field and can only access the immediate vicinity based on its filter size. There are several clues that point to this explanation. First, Table \ref{ablation_results} suggests that the attention mechanism plays a major role in the accuracy improvements. Secondly, based on Tables \ref{cifar10_results} to \ref{disco_results}, the accuracy improvements are generally higher in earlier branches, where the receptive field of the backbone network up to the branch location is lower compared to later branches. Finally, the incorporation of global scale and global information such as perspective is known to be of great importance in crowd counting, and many crowd counting methods utilize visual attention mechanisms and dilated convolution layers to this end \cite{2003.12783}, which can explain why our method performs well for this problem.

Moreover, we showed that our fine-tuning strategy, \\namely Copycat SL-ViT, has the potential to further increase the accuracy of SL-ViT branches. It is well-known that with deep learning, more data almost always improves the final outcome, and this is especially true for vision transformers which are known to be data-hungry \cite{2101.01169}. The copycat strategy can at times artificially increase the size of the dataset without introducing too much noise and thus improve the final result.

Furthermore, we introduced a novel approach for fusing audio and visual features within early exits using vision transformers. The importance of fusion inside early exits is that it creates much more options for branch locations, since a combination of any layer in the visual channel of the backbone network with any layer in the audio channel of the backbone can be selected. This allows for a more fine-grained dynamic inference, meaning a more recent result is available whenever the inference is interrupted in an anytime prediction setting, which is likely to be more accurate than earlier results.

\section*{Acknowledgment}
This work was partly funded by the European Union’s Horizon 2020 research and innovation programme under grant agreement No 957337, and by the Danish Council for Independent Research under Grant No. 9131-00119B. This publication reflects the authors views only. The European Commission and the Danish Council for Independent Research are not responsible for any use that may be made of the information it contains.

\bibliographystyle{elsarticle-num} 
\bibliography{cas-refs}

\end{document}